%% file: main.tex
\documentclass[10pt,twocolumn,letterpaper]{article}

\usepackage[pagenumbers]{main}

\usepackage[table]{xcolor}
\usepackage{microtype}
\usepackage{graphicx}
\usepackage{booktabs}
\usepackage{amsmath}
\usepackage{amssymb}
\usepackage{mathtools}
\usepackage{amsthm}

\usepackage{color}
\usepackage{enumitem}
\usepackage{multirow}
\usepackage{makecell}

\usepackage{algorithmic}
\usepackage{algorithm}
\usepackage{etoolbox,siunitx}

\definecolor{lblue}{rgb}{0.21,0.49,0.74}

\usepackage[pagebackref,breaklinks,colorlinks,citecolor=lblue]{hyperref}

\def\paperID{1}
\def\confName{RoboDrive}
\def\confYear{2024}

\title{Team Samsung-RAL: Technical Report for 2024 RoboDrive Challenge-Robust Map Segmentation Track}

\author{Xiaoshuai Hao$^{1}$, Yifan Yang$^{1}$, Hui Zhang$^{1}$, Mengchuan Wei$^{1}$, Yi Zhou$^{1}$,  Haimei Zhao$^{2}$, Jing Zhang$^{2}$\\
$^{1}$Samsung R\&D Institute China–Beijing\\$^{2}$The University of Sydney\\
{\tt\small \{xshuai.hao, yifan.yang,hui123.zhang, mc.wei, yi0813.zhou\}@samsung.com},\\ \tt\small \{h.zhao, jing.zhang1\}@sydney.edu.au}

\begin{document}

% Title
\maketitle

% Abstract
\input{sections/0_abstract}

% Main Text
\input{sections/1_intro}

\input{sections/2_related_work}

\input{sections/3_approach}

\input{sections/4_experiments}

\input{sections/5_conclusion}

% References
{
\small
\bibliographystyle{ieeenat_fullname}
\bibliography{main}
}

\end{document}

%% file: sections/0_abstract.tex
\begin{abstract}
% The ABSTRACT is to be in fully justified italicized text, at the top of the left-hand column, below the author and affiliation information.
% Use the word ``Abstract'' as the title, in 12-point Times, boldface type, centered relative to the column, initially capitalized.
% The abstract is to be in 10-point, single-spaced type.
% Leave two blank lines after the Abstract, then begin the main text.

In this report, we describe the technical details of our submission to the 2024 RoboDrive Challenge-Robust Map Segmentation Track.
The Robust Map Segmentation track focuses on the segmentation of complex driving scene elements in BEV maps under varied driving conditions.
Semantic map segmentation provides abundant and precise static environmental information crucial for autonomous driving systems' planning and navigation. 
While current methods excel in ideal circumstances, e.g., clear daytime conditions and fully functional sensors, their resilience to real-world challenges like adverse weather and sensor failures remains unclear, raising concerns about system safety.
%In this paper, we analyze and discuss the impact of different configurations on the robustness of the Map Segmentation task.
In this paper, we explored several methods to improve the robustness of the map segmentation task.
The details are as follows:
1) Robustness analysis of utilizing temporal information;
2) Robustness analysis of utilizing different backbones;
and 3) Data Augmentation to boost corruption robustness.
Based on the evaluation results, we draw several important findings including 1) The temporal fusion module is effective in improving the robustness of the map segmentation model; 2) A strong backbone is effective for improving the corruption robustness; and 3) Some data augmentation methods are effective in improving the robustness of map segmentation models.
%These findings underscore safety concerns and offer insights for designing more reliable map segmentation models.
These novel findings allowed us to achieve promising results in the 2024 RoboDrive Challenge-Robust Map Segmentation Track.

\end{abstract}

%% file: sections/1_intro.tex
%\noindent \textcolor{lblue}{\textbf{Warm greetings from the 2024 RoboDrive Challenge!}}

\section{Introduction}
\label{sec:intro}
In the rapidly evolving domain of autonomous driving, the accuracy and resilience of perception systems are paramount. 
Recent advancements, particularly in bird's eye view (BEV) representations and LiDAR sensing technologies, have significantly improved in-vehicle 3D scene perception \cite{zhang2022beverse,zhao2024simdistill}.
However, the robustness of 3D scene perception methods under varied and challenging conditions — integral to ensuring safe operations — has been insufficiently assessed. 
To fill this gap, The ICRA 2024 RoboDrive Challenge, seeks to push the frontiers of robust autonomous driving perception.
RoboDrive \cite{kong2024robodrive} is one of the first benchmarks that targeted probing the Out-of-Distribution (OOD) robustness of state-of-the-art autonomous driving perception models, centered around two mainstream topics: common corruptions and sensor failures.

Track 2 - Robust Map Segmentation requires contestants to use advanced machine learning algorithms to perform accurate map segmentation on high-resolution bird's-eye views.
This task includes a detailed analysis of various urban geographic features, such as segmented lanes, sidewalks, green spaces, etc. 
In addition, this track also tests contestants' image segmentation capabilities under different lighting, weather conditions, and noise conditions.

\section{Related Work}
\label{sec:related work}

%% file: sections/2_related_work.tex
\begin{figure*}[t]
	\setlength{\abovecaptionskip}{-0.1cm}
	\begin{center}
		\includegraphics[width=0.98\linewidth]{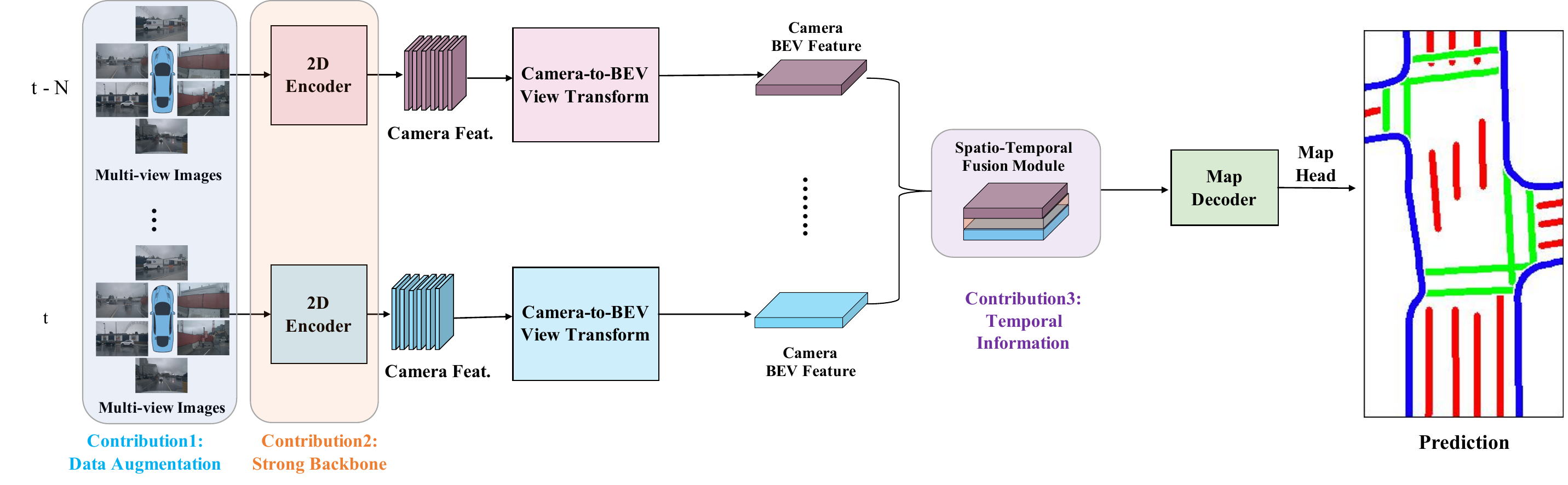}
	\end{center}
	\caption{
 An overview of Our method.
We explored several methods to improve the robustness of the baseline map segmentation model.
}
	\label{fig1}
% \vspace{-2.0em}
\end{figure*}

\begin{figure*}[t]
	\setlength{\abovecaptionskip}{-0.1cm}
	\begin{center}
		\includegraphics[width=0.9\linewidth]{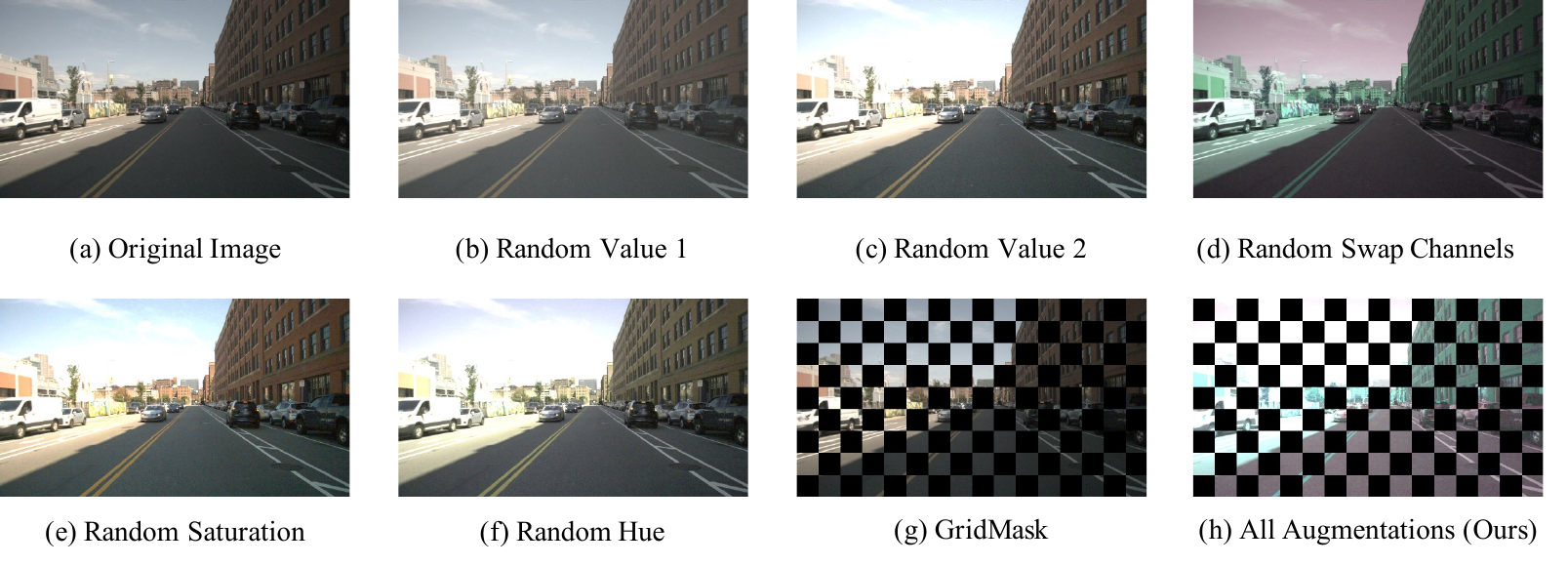}
	\end{center}
	\caption{
Examples of augmentation methods used in our method.
}
	\label{fig2}
% \vspace{-2.0em}
\end{figure*}

\subsection{Semantic Map Construction}
Semantic map construction is a prominent and extensively researched area within the field of autonomous driving.
%Recently, the design of online semantic map construction has received widespread attention.
HDMapNet~\cite{li2022hdmapnet} first introduces the problem of local semantic map learning, which aims to construct semantic
maps online from the observations of LiDAR sensors and
cameras. It also proposes a learning method to build BEV
features from sensory input and predicts vectorized map elements. BEVSegFormer~\cite{BEVSegFormer} proposes the multi-camera deformable attention to transform image-view features to
BEV representations for semantic map construction.
Very recently, BEVerse~\cite{zhang2022beverse} incorporates the semantic map construction as part of the multitask framework and uses vanilla convolutional layers for segmentation prediction. 
MBFusion \cite{haoMBFusion} proposes a multi-modal BEV feature fusion approach for the HD map construction task.
In addition, HD map construction~\cite{hao2024your,hao2024mapdistill}
have received widespread attention, which can provide fine-grained information about road scenes and play a vital role in autonomous vehicles.

\subsection{Robustness Under Natural Corruptions}
Recently, assessing autonomous driving perception model robustness under natural corruptions has burgeoned as a pivotal research domain \cite{dong2023benchmarking,zhu2023understanding,kong2023robodepth,xie2023robobev,zhao2024unimix,li2024foundation}. 
The natural corruption robustness often refers to the capability of a conventionally trained model maintaining satisfactory performance under natural distribution shifts.
ModelNet-C~\cite{ren2022benchmarking} design a taxonomy of common 3D corruptions and identify the atomic corruptions to understand classifiers' robustness.
Recently, the corruption robustness of BEV perception tasks has been widely studied.
Robodepth~\cite{kong2023robodepth,ren2022benchmarking} establishes a robustness benchmark for monocular depth estimation under corruptions.
RoboBEV~\cite{xie2023robobev} introduces a comprehensive benchmark for evaluating the robustness under natural corruptions of four BEV perception algorithms, such as 3D object detection~\cite{22eccvbevformer,liu2023bevfusion}, semantic segmentation~\cite{zhang2022beverse,zhou2022cross}, depth estimation~\cite{wei2023surrounddepth}, and semantic occupancy prediction~\cite{wei2023surroundocc,huang2023tri}.
\cite{dong2023benchmarking} systematically designs 27 types of common corruptions in 3D object detection for both LiDAR and camera sensors. Robo3D~\cite{kong2023robo3d} presents a comprehensive benchmark for probing the robustness of 3D detectors and segmentors under out-of-distribution scenarios against natural corruptions that occur in real-world environments.
%To the best of our knowledge, we are the first to study the robustness of semantic map segmentation models under sensor corruption.
In this technical report, we study the robustness of semantic map segmentation models under sensor corruption.

\begin{table*}[t]
\begin{center}
\caption{Robustness analysis of different backbones on phase 1 test set.}
\scalebox{0.7}{
  \begin{tabular}{p{3.0cm}|p{0.8cm}p{0.8cm}p{0.8cm}p{0.8cm}p{0.8cm}p{0.8cm}p{0.8cm}p{0.8cm}p{0.8cm}p{1cm}p{0.8cm}p{0.8cm}p{0.8cm}p{1cm}p{0.8cm}p{0.8cm}p{0.8cm}p{0.8cm}}
  \hline
 \rowcolor{black!10}  \makecell[c]{Model} & \makecell[c]{miou} &\makecell[c]{brigh}&\makecell[c]{dark}&\makecell[c]{fog}
 & \makecell[c]{frost} &\makecell[c]{snow}&\makecell[c]{contr}&\makecell[c]{defoc}& \makecell[c]{glass} &\makecell[c]{motio}&\makecell[c]{zoom}&\makecell[c]{elast}& \makecell[c]{gauss} &\makecell[c]{impul}&\makecell[c]{shot}&\makecell[c]{iso}&\makecell[c]{pixel}&\makecell[c]{jpeg}\\
  \midrule
  \makecell[l]{BEVerse-Tiny}
  & \makecell[c]{16.31} &\makecell[c]{16.5}&\makecell[c]{10.7}&\makecell[c]{36.4}&\makecell[c]{0.9}&\makecell[c]{4.4}&\makecell[c]{3.8}&\makecell[c]{14.7}&\makecell[c]{27.9}&\makecell[c]{16.9}&\makecell[c]{10.2}&\makecell[c]{39.3}&\makecell[c]{6.1}&\makecell[c]{4.3}&\makecell[c]{6.5}&\makecell[c]{5.6}&\makecell[c]{42.0}&\makecell[c]{32.3}\\
   
    \makecell[l]{BEVerse-Small} & \makecell[c]{17.33} &\makecell[c]{23.4}&\makecell[c]{13.5}&\makecell[c]{39.8}&\makecell[c]{3.7}&\makecell[c]{7.2}&\makecell[c]{5.7}&\makecell[c]{19.4}&\makecell[c]{29.6}&\makecell[c]{21.2}&\makecell[c]{10.5}&\makecell[c]{40.6}&\makecell[c]{3.3}&\makecell[c]{1.8}&\makecell[c]{2.8}&\makecell[c]{3.2}&\makecell[c]{40.3}&\makecell[c]{31.9}\\
   
   \rowcolor{blue!10}  \makecell[l]{BEVerse-Base}
   & \makecell[c]{24.08} &\makecell[c]{23.7}&\makecell[c]{23.7}&\makecell[c]{47.1}
   & \makecell[c]{3.8} &\makecell[c]{11.5}&\makecell[c]{7.0}&\makecell[c]{26.0}
   & \makecell[c]{39.0} &\makecell[c]{32.0}&\makecell[c]{15.5}&\makecell[c]{41.9}
   & \makecell[c]{15.0} &\makecell[c]{10.1}&\makecell[c]{14.3}&\makecell[c]{10.9}
   & \makecell[c]{48.7} &\makecell[c]{41.2}\\

  \bottomrule
  \end{tabular}}
   \label{tab-bacnbone}
   \vspace{-1em}
\end{center}
\end{table*}

\begin{table*}[t]
\begin{center}
\caption{Robustness analysis of temporal module on phase 1 test set.}
\setlength{\abovecaptionskip}{0.1em} 
\scalebox{0.65}{
  \begin{tabular}{p{4.5cm}|p{0.8cm}p{0.8cm}p{0.8cm}p{0.8cm}p{0.8cm}p{0.8cm}p{0.8cm}p{0.8cm}p{0.8cm}p{1cm}p{0.8cm}p{0.8cm}p{0.8cm}p{1cm}p{0.8cm}p{0.8cm}p{0.8cm}p{0.8cm}}
  \hline
 \rowcolor{black!10}  \makecell[c]{Model} & \makecell[c]{miou} &\makecell[c]{brigh}&\makecell[c]{dark}&\makecell[c]{fog}
 & \makecell[c]{frost} &\makecell[c]{snow}&\makecell[c]{contr}&\makecell[c]{defoc}& \makecell[c]{glass} &\makecell[c]{motio}&\makecell[c]{zoom}&\makecell[c]{elast}& \makecell[c]{gauss} &\makecell[c]{impul}&\makecell[c]{shot}&\makecell[c]{iso}&\makecell[c]{pixel}&\makecell[c]{jpeg}\\
  \midrule
  \makecell[l]{BEVerse-Base (w/o temporal)}
  & \makecell[c]{21.68} &\makecell[c]{20.4}&\makecell[c]{21.2}&\makecell[c]{40.9}&\makecell[c]{3.4}&\makecell[c]{11.0}&\makecell[c]{4.4}&\makecell[c]{24.4}&\makecell[c]{36.4}&\makecell[c]{28.3}&\makecell[c]{14.9}&\makecell[c]{39.3}&\makecell[c]{13.2}&\makecell[c]{7.6}&\makecell[c]{12.1}&\makecell[c]{9.0}&\makecell[c]{47.1}&\makecell[c]{37.3}\\

   \rowcolor{blue!10}  \makecell[l]{BEVerse-Base (w/ temporal)}
   & \makecell[c]{24.08} &\makecell[c]{23.7}&\makecell[c]{23.7}&\makecell[c]{47.1}
   & \makecell[c]{3.8} &\makecell[c]{11.5}&\makecell[c]{7.0}&\makecell[c]{26.0}
   & \makecell[c]{39.0} &\makecell[c]{32.0}&\makecell[c]{15.5}&\makecell[c]{41.9}
   & \makecell[c]{15.0} &\makecell[c]{10.1}&\makecell[c]{14.3}&\makecell[c]{10.9}
   & \makecell[c]{48.7} &\makecell[c]{41.2}\\

  \bottomrule
  \end{tabular}}
   \label{tab-temporal}
   \vspace{-1em}
\end{center}
\end{table*}

\begin{table*}[t]
\begin{center}
\caption{Robustness analysis of data augmentations on phase 2 test set.}
\setlength{\abovecaptionskip}{0.1em} 
\scalebox{0.65}{
  \begin{tabular}{p{4.5cm}|p{0.8cm}p{0.8cm}p{0.8cm}p{0.8cm}p{0.8cm}p{0.8cm}p{0.8cm}p{0.8cm}p{0.8cm}p{1cm}p{0.8cm}p{0.8cm}p{0.8cm}p{1cm}p{0.8cm}p{0.8cm}p{0.8cm}p{0.8cm}}
  \hline
 \rowcolor{black!10}  \makecell[c]{Model} & \makecell[c]{miou} &\makecell[c]{brigh}&\makecell[c]{dark}&\makecell[c]{fog}
 & \makecell[c]{frost} &\makecell[c]{snow}&\makecell[c]{contr}&\makecell[c]{defoc}& \makecell[c]{glass} &\makecell[c]{motio}&\makecell[c]{zoom}&\makecell[c]{elast}& \makecell[c]{gauss} &\makecell[c]{impul}&\makecell[c]{shot}&\makecell[c]{iso}&\makecell[c]{pixel}&\makecell[c]{jpeg}\\
  \midrule
  \makecell[l]{BEVerse-Base (w/o aug)}
  & \makecell[c]{23.24} &\makecell[c]{18.2}&\makecell[c]{28.5}&\makecell[c]{24.0}&\makecell[c]{5.4}&\makecell[c]{12.6}&\makecell[c]{5.9}&\makecell[c]{31.9}&\makecell[c]{34.1}&\makecell[c]{23.0}&\makecell[c]{21.4}&\makecell[c]{45.8}&\makecell[c]{14.6}&\makecell[c]{8.8}&\makecell[c]{11.1}&\makecell[c]{24.1}&\makecell[c]{45.5}&\makecell[c]{42.6}\\

   \rowcolor{blue!10}  \makecell[l]{BEVerse-Tiny (w/ aug)}
    & \makecell[c]{29.76} &\makecell[c]{42.4}&\makecell[c]{28.7}&\makecell[c]{30.4}
   & \makecell[c]{9.1} &\makecell[c]{8.6}&\makecell[c]{13.8}&\makecell[c]{21.6}
   & \makecell[c]{43.1} &\makecell[c]{24.8}&\makecell[c]{19.6}&\makecell[c]{75.2}
   & \makecell[c]{23.5} &\makecell[c]{14.3}&\makecell[c]{17.7}&\makecell[c]{31.4}
   & \makecell[c]{68.2} &\makecell[c]{44.5}\\

    \makecell[l]{BEVerse-Base (w/ aug)}
    & \makecell[c]{--} &\makecell[c]{--}&\makecell[c]{--}&\makecell[c]{--}
   & \makecell[c]{--} &\makecell[c]{--}&\makecell[c]{--}&\makecell[c]{--}
   & \makecell[c]{--} &\makecell[c]{--}&\makecell[c]{--}&\makecell[c]{--}
   & \makecell[c]{--} &\makecell[c]{--}&\makecell[c]{--}&\makecell[c]{--}
   & \makecell[c]{--} &\makecell[c]{--}\\

  \bottomrule
  \end{tabular}}
   \label{tab-aug1}
  \vspace{-1em}
\end{center}
\end{table*}

\begin{table*}[t]
\begin{center}
\caption{Comparisons with state-of-the-art methods.}
\scalebox{0.7}{
  \begin{tabular}{p{3.0cm}|p{0.8cm}p{0.8cm}p{0.8cm}p{0.8cm}p{0.8cm}p{0.8cm}p{0.8cm}p{0.8cm}p{0.8cm}p{1cm}p{0.8cm}p{0.8cm}p{0.8cm}p{1cm}p{0.8cm}p{0.8cm}p{0.8cm}p{0.8cm}}
  \hline
 \rowcolor{black!10}  \makecell[c]{Method} & \makecell[c]{miou} &\makecell[c]{brigh}&\makecell[c]{dark}&\makecell[c]{fog}
 & \makecell[c]{frost} &\makecell[c]{snow}&\makecell[c]{contr}&\makecell[c]{defoc}& \makecell[c]{glass} &\makecell[c]{motio}&\makecell[c]{zoom}&\makecell[c]{elast}& \makecell[c]{gauss} &\makecell[c]{impul}&\makecell[c]{shot}&\makecell[c]{iso}&\makecell[c]{pixel}&\makecell[c]{jpeg}\\
  \midrule
  \makecell[l]{Huangxl0719}
  & \makecell[c]{48.75} &\makecell[c]{54.6}&\makecell[c]{71.1}&\makecell[c]{28.6}
   & \makecell[c]{23.1} &\makecell[c]{54.5}&\makecell[c]{54.6}&\makecell[c]{64.8}
   & \makecell[c]{51.2} &\makecell[c]{44.7}&\makecell[c]{21.8}&\makecell[c]{52.1}
   & \makecell[c]{58.5} &\makecell[c]{46.1}&\makecell[c]{37.2}&\makecell[c]{64.2}
   & \makecell[c]{55.2} &\makecell[c]{54.9}\\
   
    \makecell[l]{CrazyFriday} & \makecell[c]{34.54} &\makecell[c]{42.7}&\makecell[c]{40.4}&\makecell[c]{29.8}
   & \makecell[c]{25.0} &\makecell[c]{41.7}&\makecell[c]{17.1}&\makecell[c]{42.4}
   & \makecell[c]{34.8} &\makecell[c]{33.2}&\makecell[c]{22.4}&\makecell[c]{48.0}
   & \makecell[c]{33.7} &\makecell[c]{23.6}&\makecell[c]{26.0}&\makecell[c]{38.0}
   & \makecell[c]{45.8} &\makecell[c]{45.6}\\
   
   \rowcolor{blue!10}  \makecell[l]{yf20221012 (Ours)}
   & \makecell[c]{29.76} &\makecell[c]{42.4}&\makecell[c]{28.7}&\makecell[c]{30.4}
   & \makecell[c]{9.1} &\makecell[c]{8.6}&\makecell[c]{13.8}&\makecell[c]{21.6}
   & \makecell[c]{43.1} &\makecell[c]{24.8}&\makecell[c]{19.6}&\makecell[c]{75.2}
   & \makecell[c]{23.5} &\makecell[c]{14.3}&\makecell[c]{17.7}&\makecell[c]{31.4}
   & \makecell[c]{68.2} &\makecell[c]{44.5}\\
   
  \hline
   \makecell[l]{RoboDrivers}& \makecell[c]{15.67} &\makecell[c]{21.4}&\makecell[c]{14.1}&\makecell[c]{19.2}
   & \makecell[c]{6.8} &\makecell[c]{3.2}&\makecell[c]{3.7}&\makecell[c]{18.9}
   & \makecell[c]{27.9} &\makecell[c]{9.2}&\makecell[c]{17.6}&\makecell[c]{44.6}
   & \makecell[c]{5.2} &\makecell[c]{2.1}&\makecell[c]{2.4}&\makecell[c]{9.6}
   & \makecell[c]{36.8} &\makecell[c]{28.0}\\
  \bottomrule
  \end{tabular}}
   \label{tab-main}
  \vspace{-2em}
\end{center}
\end{table*}

%% file: sections/3_approach.tex
\section{Approach}

\begin{figure*}[t]
	\setlength{\abovecaptionskip}{-0.1cm}
	\begin{center}
		\includegraphics[width=0.9\linewidth]{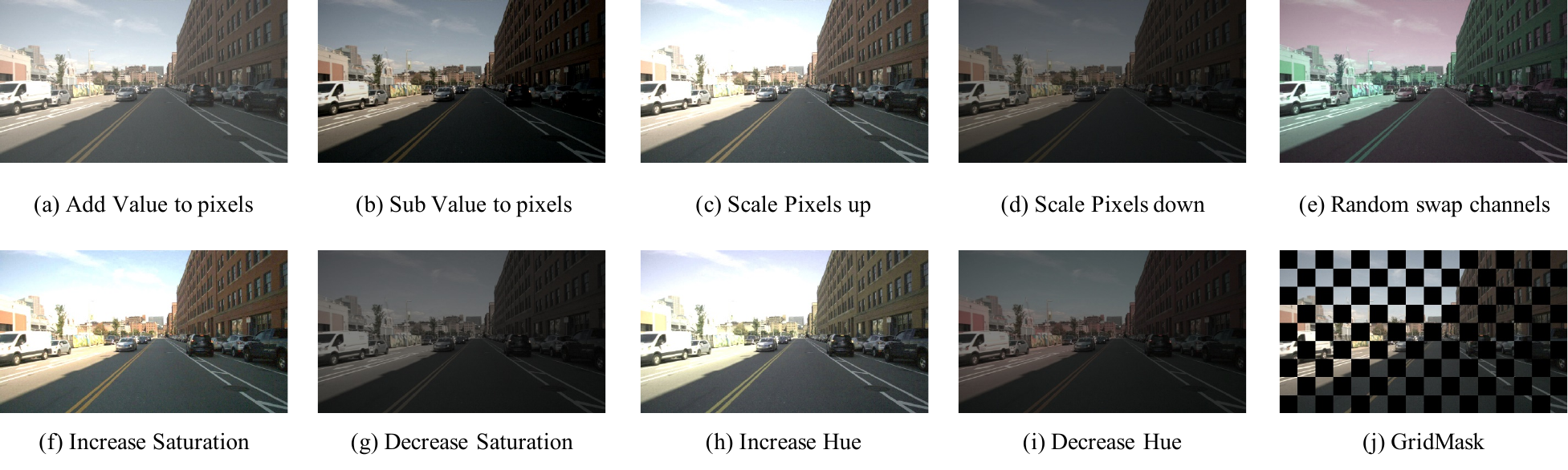}
	\end{center}
	\caption{
Illustration of different augmentation methods.
}
	\label{fig2}
% \vspace{-2.0em}
\end{figure*}

\subsection{Baseline Model}
We first establish a baseline of semantic map segmentation based on the state-of-the-art BEVerse~\cite{zhang2022beverse} model.
BEVerse takes $M$ surrounding camera images from $N$ timestamps and the corresponding ego-motions and camera parameters as input.
With multi-task inference, the outputs include the 3D bounding boxes and semantic map for the present frame, and the future instance segmentation and motion for the following $T$ frames. 
The BEVerse framework consists of four sub-modules
that are sequentially applied, including the image-view encoder, the view transformer, the spatio-temporal BEV encoder, and the multi-task decoders. 
These modules are described in detail in BEVerse~\cite{zhang2022beverse}.

%In this paper, we analyze and discuss the impact of different configurations on the robustness of the Map Segmentation task.
In this paper, we explored several methods to improve the robustness of the baseline map segmentation model.
The details are as follows:
1) Robustness analysis of utilizing temporal information;
2) Robustness analysis of utilizing different backbones;
and 3) Data Augmentation to boost corruption robustness.
The overall framework of our method is illustrated in Fig.~\ref{fig1}.
The experimental results are presented in detail in Sec.~\ref{sec4}.

%% file: sections/4_experiments.tex
\section{Experiments}
\label{sec4}

\subsection{Dataset and Implementation Details}
We implemented BEVerse~\cite{zhang2022beverse} as the baseline model for Track 2. 
The baseline model was trained on the official train split of the nuScenes~\cite{20cvprnuscense} dataset and evaluated on our robustness probing sets under different corruptions.

We train the model with 4 NVIDIA RTX A6000 GPUs. 
For training, the AdamW~\cite{loshchilov2017decoupled} optimizer is utilized, with
initial learning rate as 2e-4, weight decay as 0.01, and gradient clip as 35. 
The model is trained for 20 epochs with CBGS~\cite{zhou2019objects}. 
For the learning schedule, we apply the one-cycle
policy~\cite{yan2018second} with the peak learning rate as 1e-3. 
Moreover, for semantic map construction, the ranges
are [-30.0m, 30.0m] for X-axis and [-15.0m, 15.0m] for
Y-axis, with the interval as 0.15m.

\subsection{Metrics}
Following~\cite{li2022hdmapnet,zhang2022beverse}, the semantic classes for map construction include lane dividers, pedestrian crossings, and lane boundaries. 
For quantitative evaluation, we compute the intersection-over-union (IoU) for
each class between the predicted and ground-truth maps.
The mean IoU (mIoU) is computed as the primary evaluation metric.

\subsection{Ablation Study}
In this section, we analyze and discuss the impact of different configurations on the robustness of the Map Segmentation.

\textbf{Robustness analysis of utilizing different backbones.}
In this setting, we comprehensively investigate several backbones and show the results in Tab.~\ref{tab-bacnbone}.
%Specifically, we use three different backbones in the BEVerse~\cite{zhang2022beverse}.
We construct three versions of BEVerse, namely BEVerse-Tiny, BEVerse-Small and BEVerse-Base.
Specifically, we use Swin-Tiny, Swin-Small, and Swin-Base Transformer as the backbone, respectively.
The experimental results demonstrate that the BEVerse-Base (using a Swin-base Transformer backbone) has significantly improved the robustness.
For example, compared with BEVerse-Tiny, the BEVerse-Base achieves 7.77 absolute gains on the mIoU metric on the phase 1 test set.
This shows the effectiveness of the strong backbone in improving the corruption robustness of the semantic map segmentation model.

\textbf{Robustness analysis of utilizing temporal information.}
We investigate the impact of utilizing temporal information module~\cite{zhang2022beverse} on the robustness of semantic map segmentation and show the results in Tab.~\ref{tab-temporal}.
Two model variants, BEVerse-Base with and without the temporal fusion module, are examined.
The experimental results demonstrate that the temporal fusion module can significantly improve the robustness.
It can be observed that BEVerse-Base (with temporal fusion) achieves 2.4 absolute gains on the mIoU metric on the phase 1 test set.
This shows the effectiveness of the temporal fusion module in improving the corruption robustness of the semantic map segmentation model.

\textbf{Data Augmentation to boost corruption robustness}
In this setting, we investigate the effect of various data augmentation techniques on the robustness of map segmentation models under natural corruptions. 
Specifically, we use several image data augmentation methods, i.e., GridMask, Random Hue, Random Saturation, Random Swap Channels and Random value.
Tab.~\ref{tab-aug} lists the augmentation methods and their descriptions and ranges we used to augment data.
We choose BEVerse-Tiny (w/ temporal) as the base model, and the results are shown in Tab.~\ref{tab-aug1}.
It can be seen that image data augmentation methods moderately improve the model's robustness performance.
For example, compared with BEVerse-Base, the BEVerse-Tiny (with Aug) achieves 6.52 absolute gains on the mIoU metric on the phase 2 test set.
%For example, BEVerse-Tiny (with Aug) achieves \textbf{8.4} absolute gains on the mIoU metric.
This shows the effectiveness of the image augmentation methods in improving the corruption robustness of the map segmentation model.
\textbf{Note that} data augmentations used in our experiments are marked in gray.
In addition, we have also listed some other data augmentation methods and hope to further explore them in the future, such as Translate, Scale, Rotate, Shear, etc.

\begin{table}[!t]
    \centering
    \newcommand{\linebreakcell}{\multicolumn{2}{p{4cm}|}}
    \begin{scriptsize}
    \caption{Augmentation methods. \textbf{Data augmentations used in our experiments are marked in gray.}}
    \renewcommand{\arraystretch}{1.3} 
    \scalebox{0.95}{
    \begin{tabular}{|l|ll|l|}
        \hline
        \textbf{Name} & \linebreakcell{\textbf{Description}} & \textbf{Range} \\
            \hline
     \rowcolor{black!10}   \textit{GridMask} & \linebreakcell{Regularly drop the image information by multiplying with a grid mask, The grid mask has the size with image and its values are 0/1} &  \\

     \hline
     \rowcolor{black!10}   \textit{Random Hue} & \linebreakcell{Convert color form BGR to HSV and add a random value to image hue channel} & $[-18, 18]$ \\

        \hline
     \rowcolor{black!10}   \textit{Random Saturation} & \linebreakcell{Convert color form BGR to HSV and scale image saturation channel with rate $magnitude$} & $[0.5, 1.5]$ \\
             \hline
     \rowcolor{black!10}   \textit{Random swap channels} & \linebreakcell{ Randomly sawp the image RGB channels} &  \\

         \hline
     \rowcolor{black!10}  \textit{Random value 1} & \linebreakcell{Add a random value to all pixels} & $[-32, 32]$ \\

         \hline
      \rowcolor{black!10}  \textit{Random value 2} & \linebreakcell{Scale all pixels with rate $magnitude$} & $[0.5, 1.5]$ \\

        \hline
       \textit{Translate} & \linebreakcell{Translate the image in the horizontal and vertical direction with rate $magnitude$} & $[-0.3, 0,3]$ \\
        \hline
       \textit{Scale} & \linebreakcell{Zoom in or Zoom out the image with rate $magnitude$ and select the center of the scaled image} & $[-0.5, 0.5]$ \\
        \hline
       \textit{Rotate} & \linebreakcell{Rotate the image $magnitude$ degrees} & $[-30^{\circ}, 30^{\circ}]$ \\
        \hline
       \textit{Shear} & \linebreakcell{Shear the image along the horizontal or vertical axis with rate $magnitude$} & $[-0.3, 0.3]$ \\
        \hline
       \textit{Invert} & \linebreakcell{Invert the pixels of the image} &  \\
        \hline
        \textit{Solarize} & \linebreakcell{Invert all pixels above a threshold value of $magnitude$} & $[0, 256]$ \\
        \hline
        \textit{Equalize} & \linebreakcell{Equalize the image histogram} &  \\
        \hline
        \textit{Color balance} & \linebreakcell{Adjust the color balance of the image. A $magnitude$ = 0 gives a black and white image, while $magnitude$ = 1 gives the original image} & $[0.1, 1.9]$ \\
        \hline
        \textit{Auto contrast} & \linebreakcell{Maximize the contrast of the image} &  \\
        \hline
        \textit{Cutout} & \linebreakcell{Set a random square patch of side-length $magnitude$ pixels to gray} & $[0, 20]$ \\
        \hline
    \end{tabular} }
    \label{tab-aug}
    \end{scriptsize}
\vspace{-1em}
\end{table}

\subsection{Main Results}
For the main experiments, we use BEVerse-tiny with the temporal module as the base model and use some data augmentation, such as GridMask, Random Hue, Random Saturation, Random Swap Channels and Random value.
Tab.~\ref{tab-main} shows the overall performance of our method.
In a nutshell, our method shows significant superiority over other methods, indicating the benefit of the temporal module and data augmentations.

\textbf{Note that} due to limited time, we did not conduct related experiments on BEVerse-Base.
We believe that experimenting with strong backbone models will further improve robustness performance.

%% file: sections/5_conclusion.tex
\section{Conclusion}

In this paper, we explored several methods to improve the robustness of the map segmentation task.
By conducting large-scale experiments, we draw some important findings, as summarized below:
\begin{itemize}
\item 
By analyzing the impact of different configuration options on corruption robustness, we find that the recipes for a robust map segmentation model may include utilizing a temporal fusion module and strong backbone (\eg, Swin-Base Transformer).
\item 
Some data augmentation methods are effective in improving the robustness of map segmentation models. Nonetheless, further investigation into more advanced augmentation methods is warranted for future research.
\end{itemize}

These novel findings allowed us to achieve promising results in the 2024 RoboDrive Challenge - Robust Map Segmentation Track and offer insights for designing more reliable map segmentation models.